# Artificial Intelligence Enabled Traffic Monitoring System

Vishal Mandal[1,2], Abdul Rashid Mussah[1], Peng Jin[1], Yaw Adu-Gyamfi[1,*]

[1] Department of Civil and Environmental Engineering
University of Missouri-Columbia
E2509 Lafferre Hall, Columbia, MO 65211

[2] WSP USA
211 N Broadway Suite 2800, St. Louis, MO 63102

* Correspondence: adugyamfiy@missouri.edu

**Abstract:** Manual traffic surveillance can be a daunting task as Traffic Management Centers operate a myriad of cameras installed over a network. Injecting some level of automation could help lighten the workload of human operators performing manual surveillance and facilitate making proactive decisions which would reduce the impact of incidents and recurring congestion on roadways. This article presents a novel approach to automatically monitor real time traffic footage using deep convolutional neural networks and a stand-alone graphical user interface. The authors describe the results of research received in the process of developing models that serve as an integrated framework for an artificial intelligence enabled traffic monitoring system. The proposed system deploys several state-of-the-art deep learning algorithms to automate different traffic monitoring needs. Taking advantage of a large database of annotated video surveillance data, deep learning-based models are trained to detect queues, track stationary vehicles, and tabulate vehicle counts. A pixel-level segmentation approach is applied to detect traffic queues and predict severity. Real-time object detection algorithms coupled with different tracking systems are deployed to automatically detect stranded vehicles as well as perform vehicular counts. At each stages of development, interesting experimental results are presented to demonstrate the effectiveness of the proposed system. Overall, the results demonstrate that the proposed framework performs satisfactorily under varied conditions without being immensely impacted by environmental hazards such as blurry camera views, low illumination, rain, or snow.

**Keywords:** traffic monitoring; intelligent transportation systems; traffic queues; vehicle counts; artificial intelligence; deep learning

1. Introduction

Monitoring traffic effectively has long been one of the most important efforts in transportation engineering. Till date, most traffic monitoring centers rely on human operators to track the nature of traffic flows and oversee any incident happening on the roads. The processes involved in manual traffic condition monitoring can be challenging and time-consuming. As humans are prone to inaccuracies and subject to fatigue, the results often involve certain discrepancies. It is therefore, in best interests to develop automated traffic monitoring tools to diminishing the work load of human operators and increase the efficiency of output. Hence, it is not surprising that automatic traffic monitoring systems have been one of the most important research endeavors in intelligent transportation systems. It is worthwhile to note that most present-day traffic monitoring activity happens at the Traffic Management Centers (TMCs) through vision-based camera systems. However, most existing vision-based systems are monitored by humans which makes it difficult to accurately keep track of congestion, detect stationary vehicles whilst concurrently keeping accurate track of the vehicle count. Therefore, TMCs have been laying efforts on bringing in some levels of automation in traffic management. Automated traffic surveillance systems using Artificial Intelligence (AI) have the capability to not only manage traffic well but also monitor and access current situations that can reduce the number of road accidents. Similarly, an AI enabled system can identify each vehicle and additionally track its movement pattern characteristic to identify any dangerous driving behavior, such as erratic lane changing behavior. Another important aspect of an AI-enabled traffic monitoring system is to correctly detect any stationary vehicles on the road. Often-times, there are stationary vehicles which are left behind that impedes the flow of preceding vehicles and causes vehicles to stack-up. This results in congestion that hampers free mobility of vehicles. Intelligent traffic monitoring systems are thus, an integral component of systems needed to quickly detect and alleviate the effects of traffic congestion and human factors.

In the last few years, there has been extensive research on machine and deep learning-based traffic monitoring systems. Certain activities such as vehicle count, and traffic density estimation are limited by the process of engaging human operators and requires some artificial intelligence intervention. Traffic count studies for example require human operators to be out in the field during specific hours, or in the case of using video data, human operators are required to watch man hours of pre-recorded footage to get an accurate estimation of volume counts. This can be both cumbersome and time consuming. Similarly, when it comes to seeing traffic videos from multiple CCTV cameras, it becomes extremely difficult to analyze each traffic situation in real-time. Therefore, most TMCs seek out on deploying automated systems that can in fact, alleviate the workload of human operators and lead to effective traffic management system. At the same time, the costs associated are comparatively lower due to savings associated with not needing to store multiple hours of large video data. In this study, we deployed several state-of-the-art deep learning algorithms based on the nature of certain required traffic operations. Traditional algorithms [1-3] often record lower accuracies and fails at capturing complex patterns in a traffic scene, hence we tested and deployed deep learning-based models trained on thousands of annotated traffic images. Thus, the proposed system as shown in Figure 1 can perform the following:

1. Monitor Traffic Congestion
2. Traffic Accidents, Stationary or Stranded Vehicle Detection
3. Vehicle Detection and Count
4. Manage Traffic using a stand-alone Graphical User Interface (GUI)
5. Scale traffic monitoring to Multiple Traffic Cameras

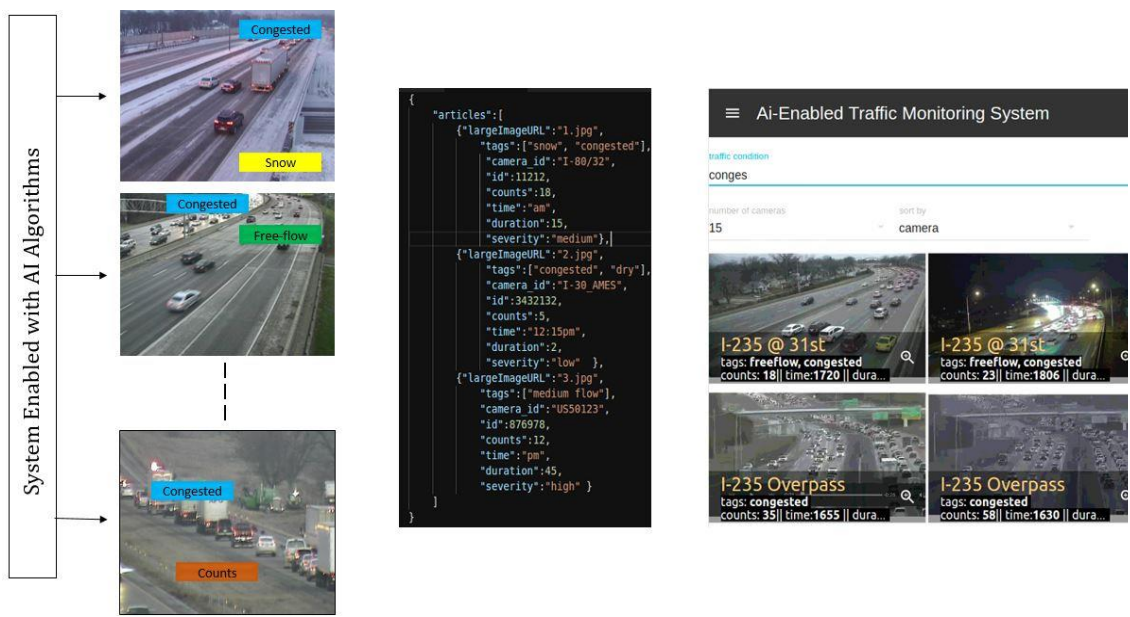

**Figure 1.** Proposed front-end GUI based system with algorithms and traffic database processed in the back-end. To visualize the demonstration of the proposed GUI based platform, refer to [4].

## 2. Literature review

In the past few years, several vision-based systems have been studied to automatically monitor traffic. We broadly discuss some of the related articles focused on congestion prediction, traffic count and anomaly detection.

*2.1 Deep Learning Frameworks for Object Detection and Classification*

There are two main ways through which video-based congestion monitoring systems function. The first instance is the "three-step-inference" based method and the other one is the "one-step-classification" based approach. Willis et al. in [5] studied traffic queues classification using deep neural networks on traffic images. The researchers trained a two-phase network using GoogLeNet and a bespoke deep subnet, and applied that in the process of detecting traffic network congestion. Chakraborty et al. in [6] used traffic imagery and applied both DCNN and YOLO algorithms in different environmental set-ups. Similarly, for inference-based approaches, Morris et al. proposed a portable system for extracting traffic queue parameters at signalized intersections from video feeds [7]. For that, they applied image processing techniques such as clustering, background subtraction, and segmentation, to identify vehicles and finally tabulated queue lengths for calibrated cameras at different intersections. Fouladgar et al. in [8] proposed a decentralized deep learning-built system wherein, every node precisely predicted each of its congestion state based on their adjacent stations in real-time conditions. Their approach was scalable and could be completely decentralized to predict the nature of traffic flows. Likewise, Ma et al. in [9] proposed an entirely automated deep neural network-based model for analyzing spatiotemporal traffic data. Their model first uses convolutional neural network to learn the spatio-temporal features. Later, a recurrent neural network is trained by utilizing the output of their first step model that helps categorize the complete sequence. The model could be feasibly applied at studying traffic flows and predicting congestion. Similarly, Wang et al. in [10] introduced a deep learning model that uses an RCNN structure to continuously predict traffic speeds. Using their model and integrating the spatio-temporal traffic information, they could identify the sources of congestion on city's ring-roads.

Popular object detection frameworks such as Mask R-CNN [11], YOLO [12], Faster R-CNN [13], etc. have been utilized far and beyond in the field of intelligent transportation systems (ITS). However, another state-of-the-art object detector called CenterNet [14] hasn't had enough exposure in ITS. So far, object detection using CenterNet has been successfully applied in the fields of robotics [15, 16], medicine [17-19], phonemes [20], etc. It's faster inference speed and smaller training time has made it popular for real-time object detection [21]. In this study, the authors deploy several state-of-the-art object detectors including CenterNet. The use of CenterNet in the of context of ITS for studying counting problem, as applied in this study, is a novel idea worth looking into, which could also further serve as literature for future studies in this area.

*2.2 Vision-based Traffic Analysis Systems*

Most existing counting methods could be generally categorized as detection instance counter [22, 23] or density estimator [24]. Detection instance counters localize every car exclusively and then count the localization. However, this could have a problem since the process requires scrutinizing the whole image pixel by pixel to generate localization. Similarly, occlusions could create another obstacle as detectors might merge overlapping objects. In contrast, density estimators work in an instinctive manner of trying to create an approximation of density for countable vehicles and then assimilating them over that dense area. Density estimators usually do not require large quantities of training data samples, but are generally constrained in application to the same scene where the training data is collected. Chiu et al. in [25] presented an automatic traffic monitoring system that implements an object segmentation algorithm capable of vehicle recognition, tracking and detection from traffic imagery. Their approach separated mobile vehicles from stationary ones using a moving object segmentation technique that uses geometric features of vehicles to classify vehicle type. Likewise, Zhuang et al. in [26] proposed a statistical method that performs a correlation-based estimation to count city's vehicles using traffic cameras. For this, they introduced two techniques, the first one using a statistical machine learning approach that is based on Gaussian models and the second one using the analytical deviation approach based on the origin-destination matrix pair. Mundhenk et al. in [27] created a dataset of overhead cars and deployed a deep neural network to classify, detect and count the number of cars. To detect and classify vehicles, they used a neural network called ResCeption. This network integrates residual learning with Inception-style layers that can detect and count the number of cars in a single look. Their approach is superior in getting accurate vehicle counts in comparison to the counts performed with localization or density estimation.

Apart from congestion detection and vehicle counts, various articles have been reviewed to study anomaly detection systems. Kamijo et al. in [28] developed a vehicle tracking algorithm based on spatio-temporal Markov random fields to detect traffic accident at intersections. The model presented in their study was capable of robustly tracking individual vehicles without their accuracies being largely affected by occlusion and clutter effects, two very common characteristics at most busy intersections which pose a problem for most models. Although traditionally, spot sensors were used primarily for incident detection [29], the scope of their use proved to be rather trivial for anomaly detection systems. Vision-based approaches have therefore been utilized far and beyond mostly due to their superior event recognition capability. Information such as traffic jams, traffic violations, accidents, etc. could be easily extracted from vision-based systems. Rojas et al. in [30] and Zeng et al. in [31] proposed techniques to detect vehicles on a highway using a static CCTV camera while, Ai et al. in [32] proposed a method to detect traffic violation at intersections. The latter's approach was put into practice on the streets of Hong Kong to detect red light runners. Thajchayapong et al. proposed an anomaly detection algorithm that could be implemented in a distributed fashion to predict and classify traffic abnormalities in different traffic scenes [33]. Similarly, Ikeda et al. in [34] used image-processing techniques to automatically detect abnormal traffic incidents. Their method could detect four different types of traffic anomalies such as detecting stopped vehicles, slow-speed vehicles, dropped objects and the vehicles that endeavored to change lanes consecutively.

## 3. Proposed Methodology

The methodology adopted for implementing an automatic traffic monitoring system is shown in Figure 2. The main components consist of first, a GPU-enabled backend (on premise) which is designed to ensure that very deep models can be trained quickly and implemented on a wide array of cameras in near real time. At the heart of the proposed AI-enabled traffic monitoring system is the development and training of several deep convolutional neural network models that are capable of detecting and classifying different objects or segmenting a traffic scene into its constituent objects. Manually annotated traffic images served as the main source of dataset used for training these models. To enable the system to be situationally aware, different object tracking algorithms are implemented to generate trajectories for each detected object on the traffic scene at all times. The preceding steps are then combined to extract different traffic flow variables (e.g. Traffic volume and occupancy) and monitor different traffic conditions such as queueing, crashes and other traffic scene anomalies. The AI-enabled traffic monitoring system is capable of tracking different classes of vehicles, tabulating their count, spotting and detecting congestion and tracking stationary vehicles in real-time.

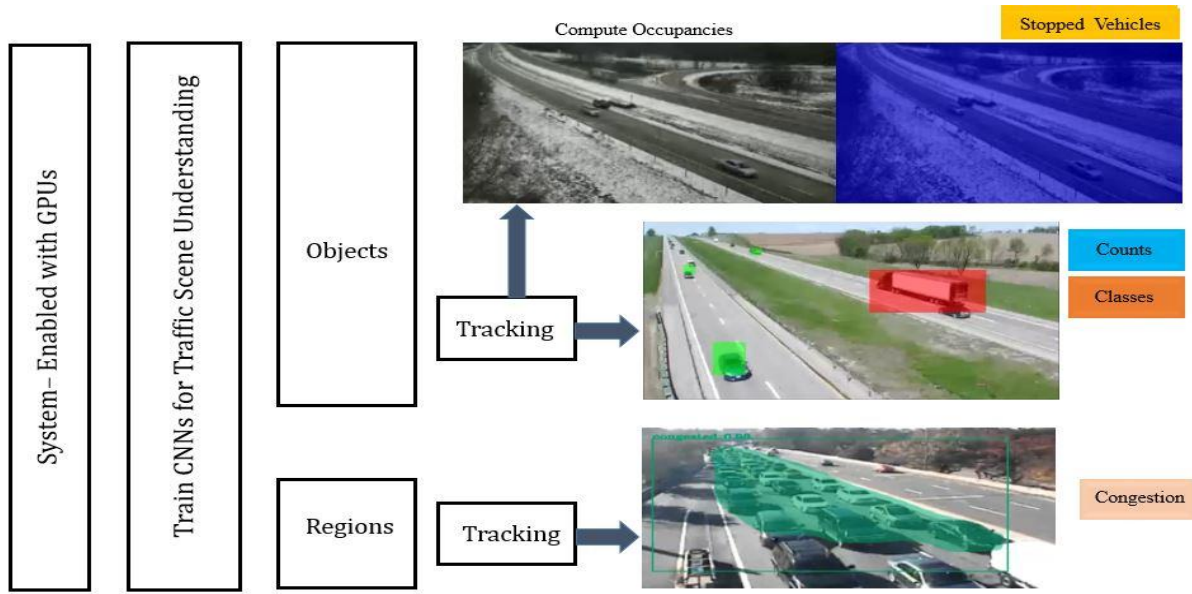

**Figure 2.** Visual Representation of the Proposed AI-Enabled System

Some of the deep learning algorithms used in the study are explained in detail as follows:

*3.1.1. Faster R-CNN*

Faster R-CNN is a two-stage target detection algorithm [13]. In Faster-RCNN, a Region Proposal Network (RPN) shares complete-image convolutional features along with a detection network that enables cost free region proposals. Here, the RPN simultaneously predicts object bounds and their equivalent score values at each position. End to end training of RPN provides high class region proposals which is used by Faster R-CNN to achieve object predictions. Compared to Fast R-CNN, Faster R-CNN produces high-quality object detection by substituting selective search method with RPN. The algorithm splits every image into multiple sections of compact areas and then passes every area over an arrangement of convolutional filters to extract high-quality feature descriptors which is then passed through a classifier. After that the classifier produces the probability of objects in each section of an image. To achieve higher prediction accuracies on traffic camera feeds, the model is trained for 5 classes viz. pedestrian, cyclist, bus, truck and car. Training took approximately 8 hours on NVIDIA GTX 1080Ti GPU. The model processed video-feeds at 5 frames per second.

*3.1.2. Mask R-CNN*

Mask R-CNN abbreviated as Mask-region based Convolutional Neural Network is an extension to Faster R-CNN [11]. In addition to accomplishing tasks equivalent to Faster R-CNN, Mask R-CNN supplements it by adding superior masks and sections the region of interest pixel-by-pixel. The model used in this study is based on Feature Pyramid Network (FPN) and is executed with resnet101 backbone. In this, ResNet101 served as the feature extractor for the model. While using FPN, there was an improvement in the standard feature extraction pyramid by the introduction of another pyramid that took higher level features from the first pyramid and consequently passed them over to subordinate layers. This enabled features at each level to obtain admission at both higher and lower-level characters. In this study, the minimum detection confidence rate was set at 90% and run at 50 validation steps. An image-centric training approach was followed in which every image was cut to the square's shape.

The images were converted from 1024×1024px×3 (RGB) to a feature map of shape 32×32×2048 on passing through the backbone network. Each of our batch had a single image per GPU and every image had altogether 200 trained Region of Interests (ROIs). Using a learning rate of 0.001 and a batch size of 1, the model was trained on NVIDIA GTX 1080Ti GPU. A constant learning rate was used during the iteration. Likewise, a weight decay of 0.0001 and a learning momentum of 0.9 was used.

The total training time for the model training on a sample dataset was approximately 3 hours. The framework for Mask-RCNN is shown in Figure 3.

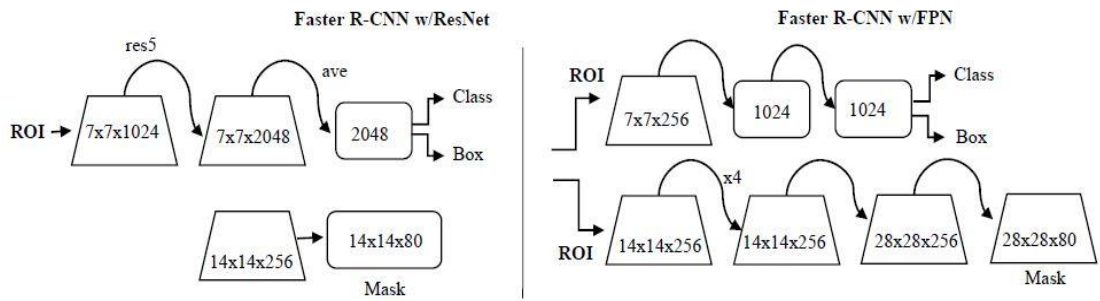

**Figure 3**. Mask R-CNN Framework

*3.1.3. YOLO*

You Only Look Once (YOLO) is the state-of-the-art object detection algorithm [12]. Unlike traditional object detection systems, YOLO investigates the image only once and detects if there are any objects in it. In this study, YOLOv4 was used to perform vehicle detection, counts, and compare results for traffic queues generation. Most contemporary object detection algorithms repurpose CNN classifiers with an aim of performing detections. For instance, to perform object detection, these algorithms use a classifier for that object and test it at varied locations and scales in the test image. However, YOLO reframes object detection i.e., instead of looking at a single image thousand times to perform detection, it just looks at the image once and performs accurate object predictions. A singe CNN concurrently predicts multiple bounding boxes and class probabilities for those generated boxes. To build YOLO models, the typical time was roughly 20-30 hours. YOLO used the same hardware resources for training as Mask R-CNN.

*3.1.4. CenterNet*

CenterNet [14] discovers visual patterns within each section of a cropped image at lower computational costs. Instead of detecting objects as a pair of key points, CenterNet detects them as a triplet thereby, increasing both precision and recall values. The framework builds up on the drawbacks encountered by CornerNet [35] which uses a pair of corner-keypoints to perform object detection. However, CornerNet fails at constructing a more global outlook of an object, which CenterNet does by having an additional keypoint to obtain a more central information of an image. CenterNet functions on the intuition that if a detected bounding box has a higher IoU with the ground-truth box, then the likelihoods of that central keypoint to be in its central region and be labelled in the same class is high. Hence, the knowledge of having a triplet instead of a pair increases CenterNet's superiority over CornerNet or any other anchor-based detection approaches. Despite using a triplet, CenterNet is still a single-stage detector but partly receives the functionalities of RoI pooling. Figure 4 shows the architecture of CenterNet where it uses a CNN backbone that performs cascade corner pooling and center pooling to yield two corner and a center keypoint heatmap. Here, cascade corner pooling enables the original corner pooling module to receive internal information whereas center pooling helps center keypoints to attain further identifiable visual pattern within objects that would enable it to perceive the central part of the region. Likewise, analogous to CornerNet, a pair of detected corners and familiar embeddings are used to predict a bounding box. Then after, the final bounding boxes are determined using the detected center keypoints.

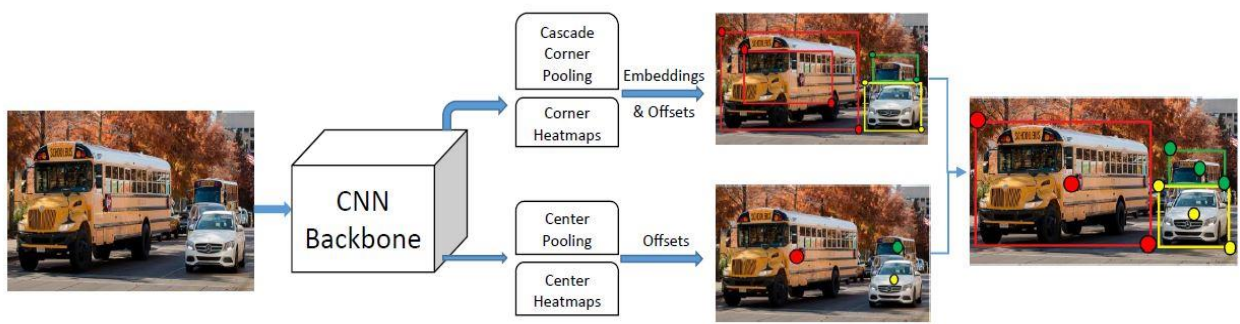

**Figure 4.** Architecture of CenterNet

The following sections give out a brief description of several traffic operations that could be seamlessly automated.

3.2 Monitoring Traffic Queues

The methodology adopted for an automatic queue monitoring system is shown in Figure 5. The first step of performing annotation was achieved using a VGG Image Annotator [36]. In the follow up, annotated images were used to train both Mask R-CNN and YOLO models. The training times for Mask R-CNN and YOLO were approximately 3.5 and 22 hours respectively. After training was done, these models were run on real-time traffic videos to evaluate their performance. The main reason for using Mask R-CNN was due to its ability to obtain pixel-level segmentation masks that made queue detections precise. Since, YOLO uses a bounding box to perform detection, it covers areas that are both congested and non-congested. Therefore, Mask-RCNN has an advantage over YOLO when it comes to precisely predicting classified regions of interest.

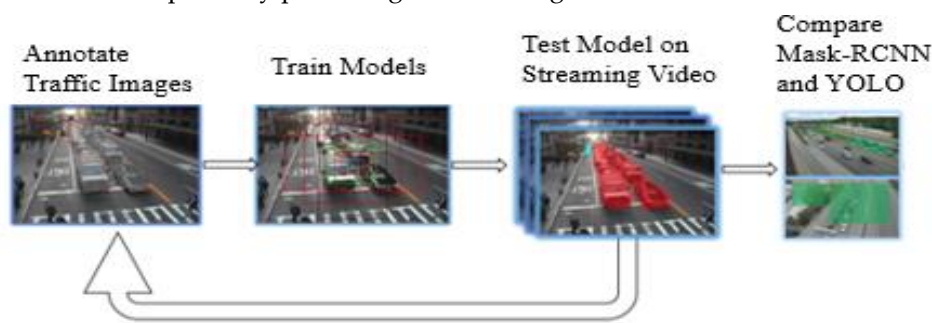

**Figure 5.** Flowchart of step-wise operations

3.3 Detecting Stationary Vehicles

Figure 6 shows the proposed methodology for detecting stationary or stranded vehicles. To begin the process, a YOLO model is trained to perform vehicle detection. Then after, detections are tracked using an Intersection over Union (IOU) process and each vehicle trajectory is plotted from traffic scenes. Tracking results are then used to sketch certain travel directions (either, east, west, north or south), the kind of road being analyzed (i.e. either intersection or freeway), and the predicted speed of tracked vehicles. The results of tracking are later used to state discrete travel directions, road type, and estimated vehicular speed. For certain types of roadway, if the vehicular speed falls under a specific threshold for a certain amount of time, then the model is able to detect that the vehicle is stationary.

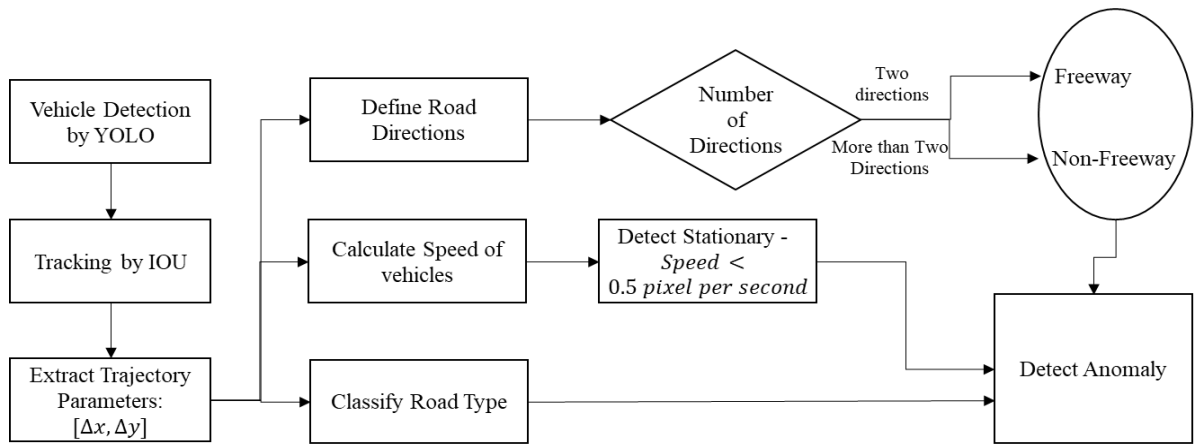

**Figure 6.** Flowchart for traffic anomaly detection

## 4. Data Description

Traffic camera images served as the primary source of dataset used in this study. The images were obtained from Iowa 511, New York State DOT, RITIS, Iowa DOT Open Data and Louisiana Department of Transportation and Development. Altogether 18,509 images were used for training and validation purposes. The datasets consisted of images taken at different times of the day in varied environmental conditions. Intersection, freeway and work-zone images were included in both training and testing samples. These images were used to train and validate deep learning models meant to carry out the processes of congestion detection, stationary vehicle tracking, and vehicle counting. For anomaly detection, traffic videos from NVIDIA AI City Challenge were used to test the effectiveness of the proposed model. Eventually, the model was assessed on 100 CCTV video feeds with different kinds of anomalies on irregular traffic and weather patterns [37].

## 5. Results

In this section, we evaluate the performance of Traffic queues, Anomaly detection system and Automatic vehicle counts.

*5.1. Traffic Queues Detection*

The performance of Mask R-CNN was tested on 1,000 traffic camera images (500 congested & 500 uncongested) and a comparative analysis is carried out with a classical YOLO framework. Standard performance metrics of precision, recall and accuracy, as shown in equations (i), (ii) and (iii) respectively were used to test the models. Then after, the results of a real-time implementation of Mask R-CNN is shown at an intersection.

$$Precision = \frac{TP}{TP\ +\ FP} \quad (i)$$

$$Recall = \frac{TP}{TP\ +\ FN} \quad (ii)$$

$$Accuracy = \frac{TP}{TP\ +\ FP+\ TN\ +\ FN} \quad (iii)$$

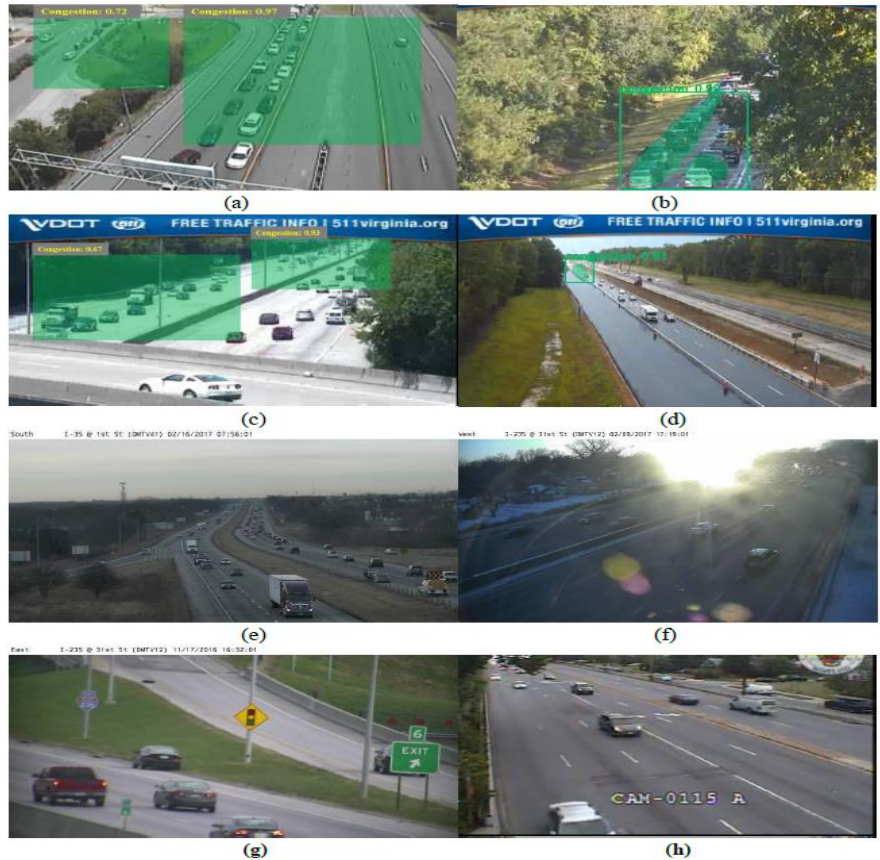

**Figure 7.** Classification of predicted queues: True Positive- (a, b), False Positive- (c, d), False Negative- (e, f), True Negative- (g, h) by YOLO & Mask R-CNN

In the Figure, above, 7(a) and 7(b) are presented as being actual predictors of congestion and are predicted as true positives (TP). Figures 7(a) and 7(b) were the detections made by YOLO and Mask-RCNN respectively. Mask R-CNN predicts congestion using a pixel-wise segmentation method while YOLO predicts congestion through a bounding box approach. Similarly, Figures 7(c) and 7(d) were the misclassifications and thus, classified as false positives (FP). In Figure 7(d), Mask R-CNN incorrectly predicted an uncongested image as congested due to the presence of a distant platoon of vehicles present in the image. The existence of an overhead bridge might have caused YOLO to make a mistake in Figure 7(c). Example of false negatives (FN) are shown in Figure 7(e) and 7(f) where both Mask R-CNN and YOLO were unsuccessful at detecting queues. The reason could be that the platoon of vehicles was far away from the camera (Figure 7e) and glaring issues (Figure 7f). Figure 7(g) and 7(h) correctly predicted uncongested images as true negatives (TN).

5.1.1. A case study for studying Traffic Queues

A case study was conducted where the Mask RCNN model was implemented in real time for monitoring queues at an intersection. It is imperative to note that the alterations in video camera perspective often made it challenging to extract traffic queue parameters from frame scenes. A typical course around this was to adjust the camera to a specific height, observing angle, zoom level, etc. Though this might be effective but is not scalable. Another alternative to this approach could be to directly use image pixel values to characterize queue parameters. While using this method, queue information from one spot could not be compared to a different location since the camera geometric features could possibly differ. In the steps described below we develop a simple, standardized calibration free approach for extracting queue length parameters from traffic video feeds. This approach is scalable and is useful in comparing queuing levels at different locations.

Step 1: Extract queue regions from traffic video-feeds with Mask RCNN.
Step 2: Calculate the pixel length of each detected queue mask.
Step 3: Accumulate length over time (minimum duration is 1 week).
Step 4: Use adaptive thresholding (Figure 8) to bin queue lengths into different severity levels: low, medium and high.
Step 5: Generate heat map of queuing levels and finally, compare.

---

Steps shown for Adaptive Thresholding

---

Initialize: L, M, H

Input: PL – pixel lengths

**for** each location **do**

    **for** each [day, hour, minute] in [30 days, 24 hours, 60 minutes] **do**

        *% extract first, second, third quartile pixel lengths*

        $Q = \text{percentile}[PL, \{Q1, Q2, Q3\}]$

    **end**

    $L = Q[\{Q1, Q2, Q3\}].\text{mean}.\text{max} + k * Q[\{Q1\}].\text{std}$

    $M = Q[\{Q1, Q2, Q3\}].\text{mean}.\text{max} + k * Q[\{Q2\}].\text{std}$

    $H = Q[\{Q1, Q2, Q3\}].\text{mean}.\text{max} + k * Q[\{Q3\}].\text{std}$

**end**

**Output:** L, M, H

**Figure 8.** Adaptive Thresholding Steps

The Mask R-CNN framework was used to quantify queuing levels at an intersection. The heat map shown in Figure 9 clearly captures the onset and dissipation of queues. The heat map for the intersection could detect both AM and PM peak hours.

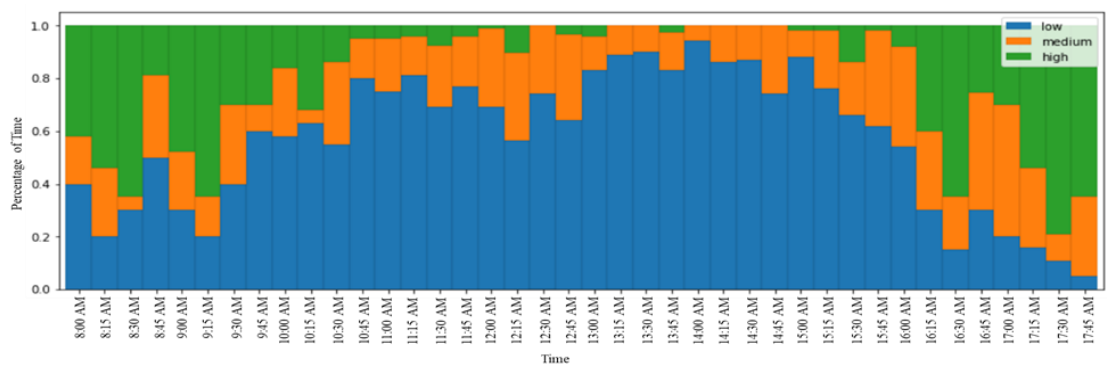

**Figure 9.** Heat map of traffic queue severity at an intersection

*5.2. Stationary Vehicle Detection*

Faster R-CNN and YOLO algorithms were deployed to study stationary vehicles. To comprehend and compare the test results for both Faster R-CNN and YOLO frameworks, confusion matrices and F-1 scores were used. The confusion matrix represents accuracy levels for different sections of image classification. Overall, 25 test results are shown in a 5 x 5 table that is referred to as a confusion matrix. Here, each row shows the actual number of predictions and total number of each row implies the number of targets predicted for that class. Likewise, every column signifies the true number of targets while the total number of each column represents the actual number of targets for that class. Similarly, F-1 score shown in equation (iv) is used to compare the performance of both Faster R-CNN and YOLO models. The results obtained for confusion matrix and F-1 scores are shown in Tables 1 and 2.

$$F-1 = \frac{2 \times Precision \times Recall}{Precision + Recall} \quad (iv)$$

As seen from Table 1, the performance of both Faster R-CNN and YOLO models were similar. Faster R-CNN was relatively inferior in detecting cyclist and bus but was better at detecting trucks when compared to the performance of YOLO. Both models predicted cars and pedestrians with a 99% level of accuracy. From Table 2, it is understood that the cumulative F-1 score of YOLO was lower than that of Faster R-CNN. Also, the recall value for YOLO was lower which implies that YOLO detects fewer objects on a traffic scene compared to Faster R-CNN. After comparing results in Table 2, it appears that Faster-RCNN was slightly better but comparable to YOLO. Therefore, any one of them could be used as an object detector.

**Table 1.** Confusion Matrix of YOLO & Faster R-CNN

| YOLO | | | | | |
|---|---|---|---|---|---|
| Pred / True | **Ped** | **Cyclist** | **Car** | **Bus** | **Truck** |
| **Ped** | 0.9928952 | 0.0053286 | 0.000888099 | 0 | 0.000888099 |
| **Cyclist** | 0.02283105 | 0.97260274 | 0 | 0 | 0.00456621 |
| **Car** | 0 | 0 | 0.994734751 | 0.000198689 | 0.005066561 |
| **Bus** | 0 | 0 | 0 | 0.994285714 | 0.005714286 |
| **Truck** | 0 | 0 | 0.045793397 | 0.007454739 | 0.946751864 |
| RCNN | | | | | |
| Pred / True | **Ped** | **Cyclist** | **Car** | **Bus** | **Truck** |
| **Ped** | 0.99737073 | 0.00262927 | 0 | 0 | 0 |
| **Cyclist** | 0.04017857 | 0.95535714 | 0.004464286 | 0 | 0 |
| **Car** | 0.00029163 | 0.00019442 | 0.994361816 | 0.00038884 | 0.004763293 |
| **Bus** | 0 | 0 | 0.010362694 | 0.979274611 | 0.010362694 |
| **Truck** | 0 | 0 | 0.0367428 | 0.007944389 | 0.95531281 |

**Table 2.** F-1 Scores of YOLO & Faster R-CNN

| **YOLO** | | | |
|---|---|---|---|
| **Class** | **Precision** | **Recall** | **F-1 score** |
| Ped | 0.92165122 | 0.7367003 | 0.81886228 |
| Cyclist | 0.94247788 | 0.8658537 | 0.90254237 |
| Car | 0.92769723 | 0.7990594 | 0.85858674 |
| Bus | 0.95081967 | 0.8571429 | 0.9015544 |
| Truck | 0.91603875 | 0.840079 | 0.87641607 |
| **Total** | 0.92690348 | 0.7975653 | 0.85738408 |
| **Faster R-CNN** | | | |
| Ped | 0.87549251 | 0.88385044 | 0.87965162 |
| Cyclist | 0.9380531 | 0.85140562 | 0.89263158 |
| Car | 0.83125773 | 0.87880373 | 0.85436976 |
| Bus | 0.8952381 | 0.86635945 | 0.88056206 |
| Truck | 0.89282203 | 0.8972332 | 0.89502218 |
| **Total** | 0.84178622 | 0.87989299 | 0.86041789 |

Similarly, after the object detector spots any vehicle's position on a traffic scene, the tracker is brought in to track the state of vehicles from a sequence of traffic video frames. Intersection over Union (IOU) and feature-based tracking systems have been deployed and further explained as follows:

*5.2.1. Tracking Detection by IOU and Feature Tracker*

Anomaly detection systems not only require detector to correctly detect vehicles in the frames, but also need tracker to distinguish the state and motion of each vehicle. After the detector predicts the position of vehicle in each frame, the tracker is liable for tracing vehicle trajectory based on a series of consecutive frames within a video file. After calculating the spatial overlap of object detection boxes in each consecutive video frame, an IOU allocates detections. Erik et al.'s IOU was implemented in this study [38]. As IOU trackers have lower computational cost, obtaining trajectories of vehicles is easy to attain and integrate to other higher-level trackers without affecting the computational speed. Frame rates even as high as 50,000 fps can be achieved with IOU. It is imperative to note that the IOU tracker is heavily reliant on how accurately predictions are done by object detection models. Road type is categorized based on the number of street directions detected. For more than two detected directions, the road type is categorized as either an intersection or an interchange. Likewise, for exactly two detected directions, the road is categorized as a freeway or simply, a two-lane street. In Figure 10, the first image is classified as a freeway while the second image is an intersection.

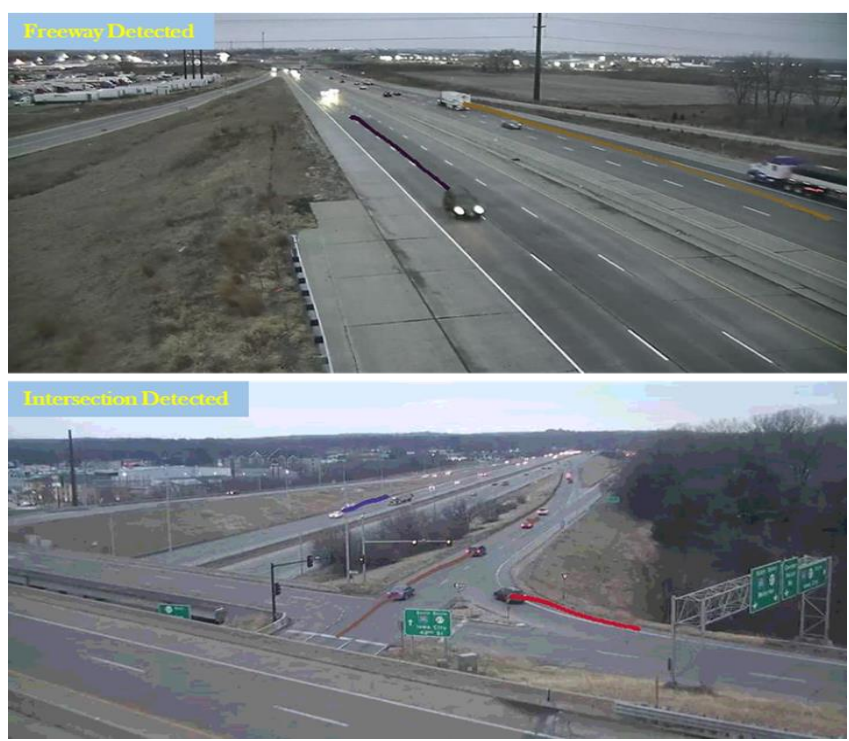

**Figure 10.** Vehicle tracking and road type classification

In Feature-based object tracking, appearance information is used to track objects in a traffic scene. This method is useful in tracking vehicles in an environment where occlusion frequently occurs. The system extracts object features from one frame and then matches appearance information with succeeding frames based on the level of similarity. The minimum value of cosine distance is suitable for calculating any resemblance between some of the characteristic features which is convenient for vehicle tracking. Besides, the results are compared between IOU and Feature Tracker based on the average switch rate for different environmental and video quality conditions. The switch rate measures how commonly a vehicle is assigned a new track number when it crosses a traffic scene. In simple words, it is the ratio of vehicle switch to the actual number of vehicles.

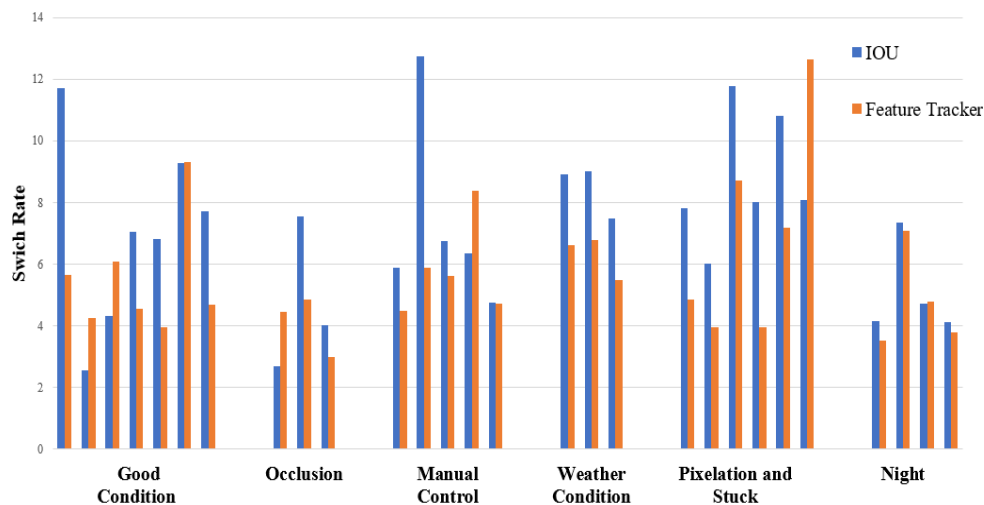

**Figure 11.** Comparison of Clustered Charts for IOU and Feature Tracker

In this study, an anomaly is defined as an event whereby any vehicle stops for 15 seconds or more, typically in a non-congested environment. To detect anomalies, the speed of every tracked vehicle is calculated over time. Based on that, any vehicle beyond the speed of 0.5 pixels per second over a 15 second time interval is characterized as a probable anomaly. Likewise, the direction of travel and the type of road is used to decide the possibility of anomaly in post-processing steps. The detected traffic anomalies are shown in Figure 12. These anomalies are shown both prior to and after post-processing of the required steps. The impact of ID switches from the IOU tracker is fairly apparent in the second column of Figure 11. This, in fact causes several anomalies to be detected at the same spot. In the post-processing step, an ID suppressing technique is used to decrease the number of anomalies. In order to achieve this, the first step is to detect multiple anomalies that remain close to one another which are then combined to one. After that, all the anomalies are merged based on the direction of roadway. The assumption made here is that only one anomaly exists on one side of the road within a 15-minute time interval. Finally, traffic anomalies are plotted in case roadway is either a freeway or a two-lane street and if the road is assessed as an intersection, then the anomaly is rejected and considered a false case.

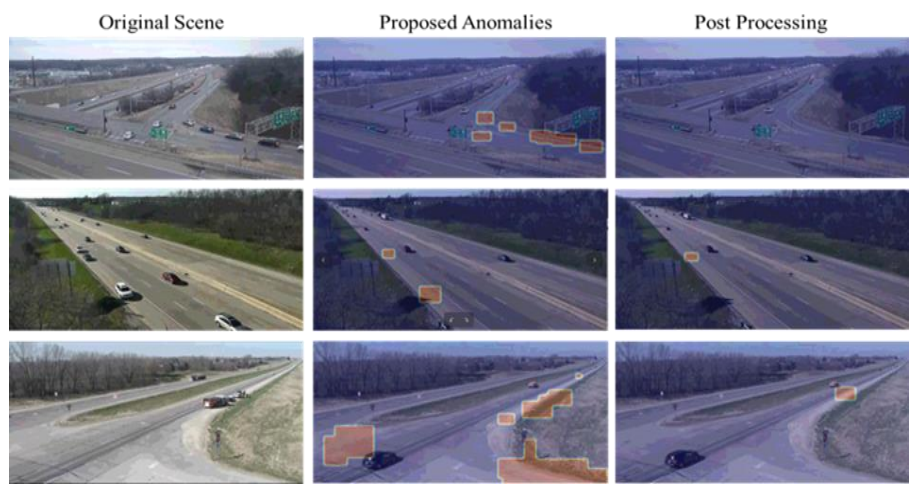

**Figure 12.** Traffic Anomalies

The proposed traffic anomaly detection system was assessed on 100 traffic video-feeds with varying traffic and weather patterns. The presence of frozen frames and pixilation effects in the assessment video dataset presented a major challenge in detecting anomalies. The IOU tracker used in the study conceived a single vehicle or a platoon of vehicles a possible anomaly even if the traffic stop sign dictated them to stop. Although, this condition could not be classified as an anomaly, the IOU labelled them as such. Therefore, to overhaul this issue, it is important to determine whether a roadway is an intersection or a freeway. Based on the road type, any vehicle remaining stationary for over 30 seconds on a freeway was considered an anomaly and for an intersection, the time limit was set to 60 seconds. Similarly, IOU tracker's competence was further challenged by video files that remained stuck for certain time periods. The videos often remained stuck for over a minute. In such cases, the IOU tracker detected the vehicle as a probable anomaly each time the video was frozen for longer time periods. This could however, be classified as a false anomaly. While conducting the experiment, it was identified that although, the video remained frozen for longer time periods, the speed of each vehicle in the frozen video remained 0, as it's the same video-frame scene. Since, any vehicle's speed in an accident is approximately 0 although not exactly zero, the rectangle surrounding it is in somewhat swaying state. Therefore, all anomalies with a speed value of zero were categorized as false detections. To determine the performance of the proposed anomaly detection model, standard performance metrics of F1, Root Mean Square Error (RMSE) and S3 values were used. The equation used to compute the value of S3 score is shown in equation (v).

$$S3 = F1 * (1 - NRMSE) \qquad (v)$$

As shown in equation (v), NRMSE is the Normalized Root Mean Square Error. To compute the F-1 score, the value for True Positive is required. A true positive is defined as the one in which the detection of an anomaly is under the 10 seconds' time frame from the actual. An anomaly can only be considered a true positive for a single anomaly. In other words, the same anomaly could not be counted twice. False positive cases are defined as ones that do not resemble to true positives for certain occurrences. Similarly, false negatives are the type of anomalies that are true anomalies in nature but are missed by the model. Figure 13 shows some of the examples for True Positive, False Positive and False Negative.

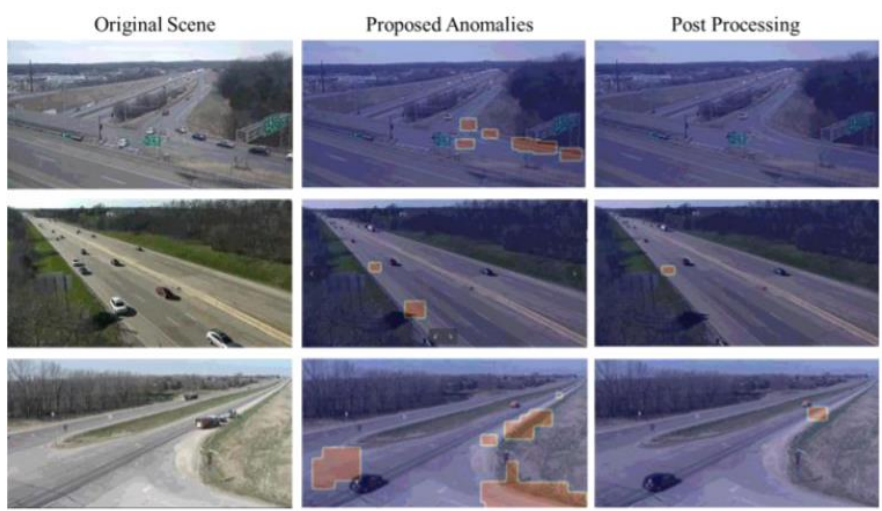

**Figure 13.** Classification of predicted anomalies - First Row, (a): True Positives. Second Row, (b): False Negatives - anomalies indicated with red circles. Third Row, (c): False Positives.

Errors in anomaly detection are represented by the root mean square error (RMSE). The RMSE value is calculated for the ground truth anomaly times and predicted anomaly times for any true positive's detections. S3 value is computed using the RMSE, that is normalized by NRMSE as seen from equation (v). Normalization is carried out using a min-max normalization technique with the largest and lowest values set at 300 and 0 respectively. From Table 3, the F1 score is calculated to be 0.8333, meaning that the detector predicts nearly 83.3% of the total anomalies. However, due to the shortcomings in the dataset, specifically for vehicles situated distant from the camera, the model failed to spot anomalies in those situations.

Table 3. F1, RMSE and S3 Final values

| Name | F1 | RMSE | S3 |
| --- | --- | --- | --- |
| Model M1 | 0.8333 | 154.7741 | 0.4034 |

The importance of anomaly detection algorithms extends the use-case beyond not only detecting traffic incidents in real-time, but also being able to properly and accurately measure and calculate their durations and secondary effects of such incidents, be it either queue formations or the possibility of secondary downstream incidents of the formed queue. It is no surprise to know that traffic incidents account for a quarter of all roadway congestion in the United States [39]. Average clearance time for incidents reported through the HELP [40] program ranged between 42 minutes to 50 minutes. The usual approach to measuring the impact of traffic incidents utilizes deterministic queuing diagrams, coupled with an examination of the change of network capacity [41]. A challenge in achieving effective incident management is due to the lack of accurate data that quantifies the impact of incidents, taking into account both their unique spatial and temporal attributes [42]. Traffic incident management response can benefit from the valuable insights extrapolated from the data derivable from detected incident situations, as well as the effects of the applied countermeasures, in order to improve on secondary responder deployment and coordination to the benefit and improvement of future situation management.

*5.3. Vehicle Counts*

With the advent of ITS, vehicle counts are often automated using either loop detectors or vision based systems. Although inductive loops give out accurate traffic counts, they often have trouble distinguishing the type of vehicles (i.e. cars, trucks, buses, etc.). Not to forget that these detectors are intrusive. On the contrary, vision based systems' non-intrusive nature enables counts by different vehicle class types with high confidence scores [43, 44]. Since, accurate vehicle count enable TMCs and other transportation agencies apply them in their day to day application areas, the significance of accurate vehicle counts cannot be ignored. Studies such as daily volume counts, travel times calculation, and traffic forecasts are all precursors of an accurate vehicle counting system. These parameters serve as important tools for optimizing traffic at different roadways. Similarly, counting information also enable engineers to obtain future traffic forecasts which in turn helps identify what routes are utilized extensively to lay out affirmative planning decisions.

In this study, we aim at developing a single look vehicle counting system that could automatically detect and tabulate the number of vehicles passing through the road. To accurately perform vehicular count, the vehicles are detected using object detectors and then traced through trackers. To obtain vehicle counts, the trackers are set an IOU threshold of 0.5 as shown in equation (vi) which helps correctly track vehicles and avoids multiple counts.

$$IOU = \frac{Intersection}{Union} \qquad (vi)$$

To assess the performance of the proposed models, the number of vehicles passing through the north and southbound directions were manually counted and compared against the automatic counts obtained from the combination of two different object detectors and trackers. CenterNet and YOLOv4 were the two different object detectors used in combination with IOU and Feature Tracker. For comparison, these frameworks were tested on a total of 546 video clips each 1 minute in length comprising of over 9 hours of total video length.

**Table 4.** Vehicle Count Performance

| Time of Day | Detector/Tracker Combination | Northbound Count Percentage | Southbound Count Percentage |
| --- | --- | --- | --- |
| Day | CenterNet and IOU | 137.04 | 144.06 |
| | CenterNet and Feature Tracker | 75.02 | 105.66 |
| | YOLOv4 and IOU | 144.38 | 155.27 |
| | YOLOv4 and Feature Tracker | 70.81 | 89.70 |
| Night | CenterNet and IOU | 144.75 | 161.38 |
| | CenterNet and Feature Tracker | 74.74 | 112.41 |
| | YOLOv4 and IOU | 145.91 | 166.23 |
| | YOLOv4 and Feature Tracker | 72.99 | 87.12 |
| Rain | CenterNet and IOU | 169.74 | 150.31 |
| | CenterNet and Feature Tracker | 119.14 | 99.47 |
| | YOLOv4 and IOU | 145.91 | 153.76 |
| | YOLOv4 and Feature Tracker | 82.06 | 74.89 |

Table 4 demonstrates the performance comparison of CenterNet and YOLOv4 models in different conditions. The performance of these detector-tracker frameworks is assessed by dividing the values obtained from them with the manually counted ground truths expressed in per hundredth or percentage. As seen from Table 4, the combination of YOLOv4 and Feature tracker obtained a reasonable counting performance for all the three different environmental conditions specified. For model combinations where a count percentage of over one hundred was achieved, there was clearly some fault in both detector and tracker. The reasoning behind the detector-tracker combination achieving over 100 percent accuracy is largely to do with the object detector generating multiple bounding boxes for the same vehicle. This resulted in overcounting of vehicles. Similarly, IOU at times did not do very well at predicting vehicle trajectories and identified them as disparate vehicles.

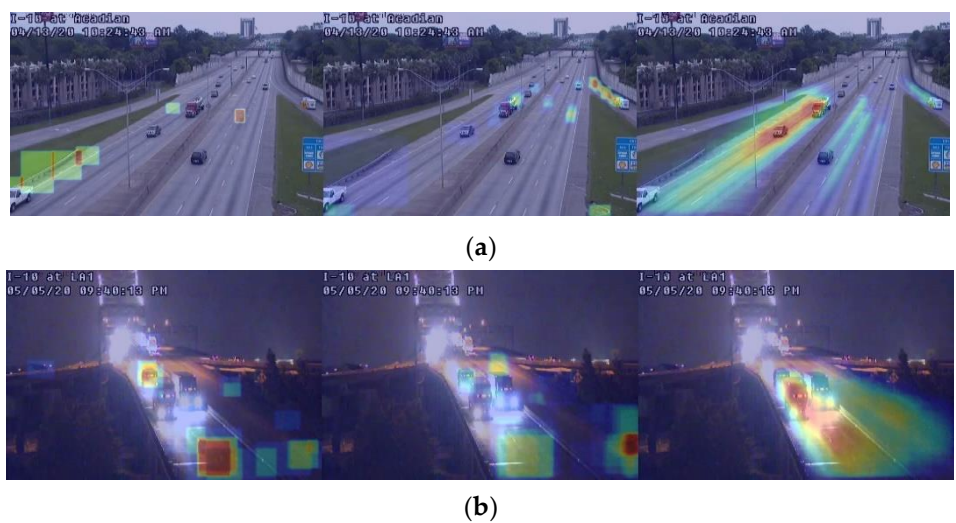

(**a**)

(**b**)

**Figure 14.** Heat Maps generated for vehicle counts using (a) CenterNet and (b) YOLOv4

To study the performance of object detectors, heat maps showing False Negatives (FN), False Positives (FP) and True Positives (TP) from left to right, are shown in Figure 14 for CenterNet and YOLOv4 models. YOLOv4 did well at detecting FN however, CenterNet detected multiple vehicles as seen from the generated heat maps in its south bound direction. This was largely due to the insufficient number of traffic images used for training. Another possibility is that the model experienced heavy congestion at these locations due to the presence of heavy gross-weighted vehicles such as buses and trucks. The FP for object detectors are generally clean for both the object detectors which is ideal for this situation. Some instances of FP could be seen from YOLOv4 which could have resulted due to the lower visibility or night-time conditions. For TP, both CenterNet and YOLOv4 models generated accurate predictions with an exclusion of a specific situation where the vehicles were too distant from the camera.

**6. Front-End Graphical User Interface**

React, [45] a JavaScript library, was used to build a front-end Graphical User Interface (GUI). The deep learning algorithms are made to run in the background on live-traffic video feeds. These algorithms record the state of traffic flows such as congestion, environmental conditions (i.e. rain, snow), etc. and display CCTVs for roadways on their front-end GUI with their constituent levels of traffic severity just by type writing certain keywords. For example, a traffic operator at the TMC wants to know what camera locations spot congestion or estimate the number of vehicles on that section of the roadway. The operator can just do that by merely typing a bunch of keywords on the GUI's input panel and the system would display the list of cameras that record congestion. Similarly, additional information such as vehicle counts on the camera locations help operators extrapolate traffic density information at certain times of the day at that location. Factors such as weather information also exert a great sense of importance for studying traffic behavior. Out of many other functionalities, the proposed system also enables the operator to identify what camera locations observe different weather patterns such as if there's rain or snow right that moment in time. For instance, how would the traffic need to be managed in situations where recurring congestion occur due to weather impacts such as heavy rainfall or snow storms. All this information serve as useful tools in discerning appropriate traffic monitoring needs by quickly running over hundreds of cameras and enabling operators ease and accessibility in traffic surveillance. For further detail, please refer to [4] to see a quick demonstration of the developed GUI.

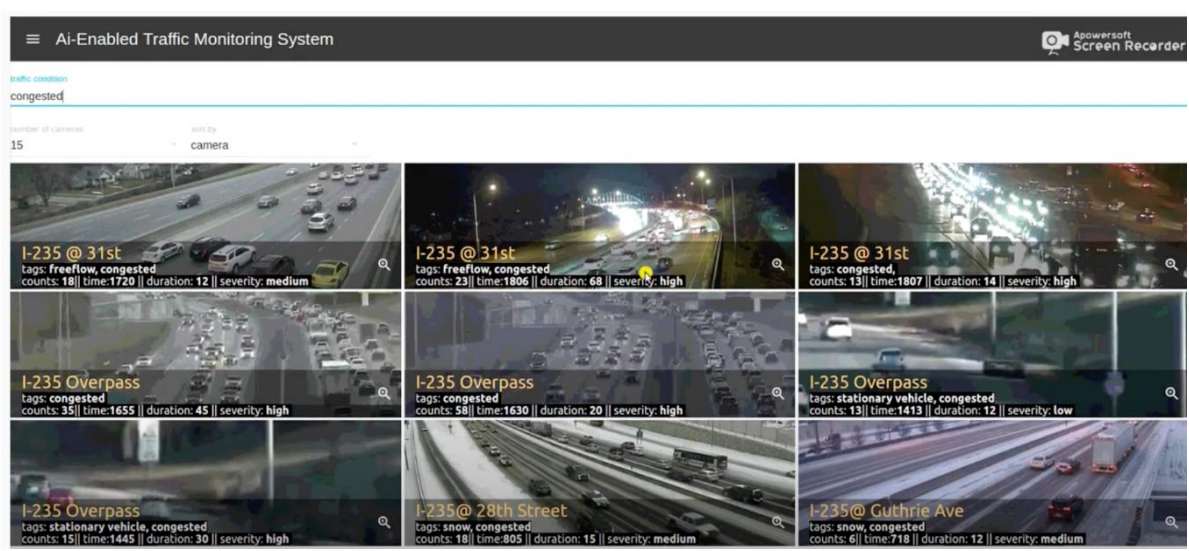

**Figure 15.** Screenshot of AI-Enabled Traffic Monitoring System GUI

## 7. Conclusion

The rapid progression in the field of deep learning and high-performance computing has highly augmented the scope of video-based traffic monitoring systems. In this study, an automatic traffic monitoring system is developed that builds up on robust deep learning models and facilitates traffic monitoring using a graphical user interface. Deep learning algorithms, such as Mask R-CNN, Faster R-CNN, YOLO and CenterNet were implemented alongside two different object tracking systems viz. IOU and Feature Tracker. Mask R-CNN was used to detect traffic queues from real-time traffic CCTVs whereas YOLO and Faster R-CNN were deployed to predict objects which later coupled with object trackers were used to detect stationary vehicles. Mask R-CNN predicted traffic queues with 92.8% accuracy while the highest accuracy attained by YOLO was 95.5%. The discrepancy in correctly detecting queues was mainly due to the poor image quality, queues being distant from the camera and glaring effects. These issues significantly affected the accuracies of the proposed models. Similarly, the F1, RMSE and S3 scores for detecting stationary vehicles were 0.8333, 154.7741, and 0.4034 respectively. It was observed that the model correctly detected stranded vehicles which remained closer to the camera but faced difficulties while detecting distant stationary vehicles. Part of the problem for lower S3 scores was also due to issues such as video pixelation, and the presence of traffic intersections. Regardless, procedures such as anomaly suppression and video pixelation corrections were useful at improving the efficacy of the proposed model. It is worthwhile to note that these corrections led to an effective stationary vehicle prediction system. Lastly, the performance of vehicle counting framework was satisfactory for both CenterNet and YOLO' combinations with Feature Tracker. However, the vehicle counting framework could be further explored and the existing models be further fine-tuned to generate a near to perfect counting framework. This could in fact be ideal for most transportation agencies as they rely heavily on turning movement counts to optimize traffic signals at intersections.

In conclusion, the proposed models which form an integrated AI enabled traffic monitoring system obtained superior results and could be useful at attaining some level of automation at Traffic Management Centers. It is worth mentioning that since, most software suites sold by transportation vendor companies cost over hundreds of thousands of dollars, their functionalities are limited; and offer just a few additional traffic surveillance capabilities than our proposed framework. In that case, the system proposed in this paper could be a cheaper and reliable alternative to bringing in some level of traffic automation by supplementing it with some additional low-cost back-up software suites.